\documentclass{article} 
\usepackage[final]{corl_2018} 
\usepackage{graphicx}
\usepackage{subcaption}
\usepackage{pgfplots}
\pgfplotsset{compat=newest} 
\usepgfplotslibrary{units} 

\title{Experiential Robot Learning with Accelerated Neuroevolution}

\author{
Ahmed Aly\\
Electrical and Computer Engineering Department \\
University of Virginia \\
Charlottesville, VA 22904, USA \\
\texttt{aaa2cn@virginia.edu} \\
\And
Joanne B. Dugan\\
Electrical and Computer Engineering Department \\
University of Virginia \\
Charlottesville, VA 22904, USA \\
\texttt{jbd@virginia.edu} \\
}

\begin{document}
\maketitle

\begin{abstract}
Derivative-based optimization techniques such as Stochastic Gradient Descent has been wildly successful in training deep neural networks. However, it has constraints such as end-to-end network differentiability. As an alternative, we present the Accelerated Neuroevolution algorithm. The new algorithm is aimed towards physical robotic learning tasks following the Experiential Robot Learning method. We test our algorithm first on a simulated task of playing the game Flappy Bird, then on a physical NAO robot in a static Object Centering task. The agents successfully navigate the given tasks, in a relatively low number of generations. Based on our results, we propose to use the algorithm in more complex tasks.
\end{abstract}


\section{Background}
\begin{figure}[b]
\begin{center}
\includegraphics[width=0.3\linewidth]{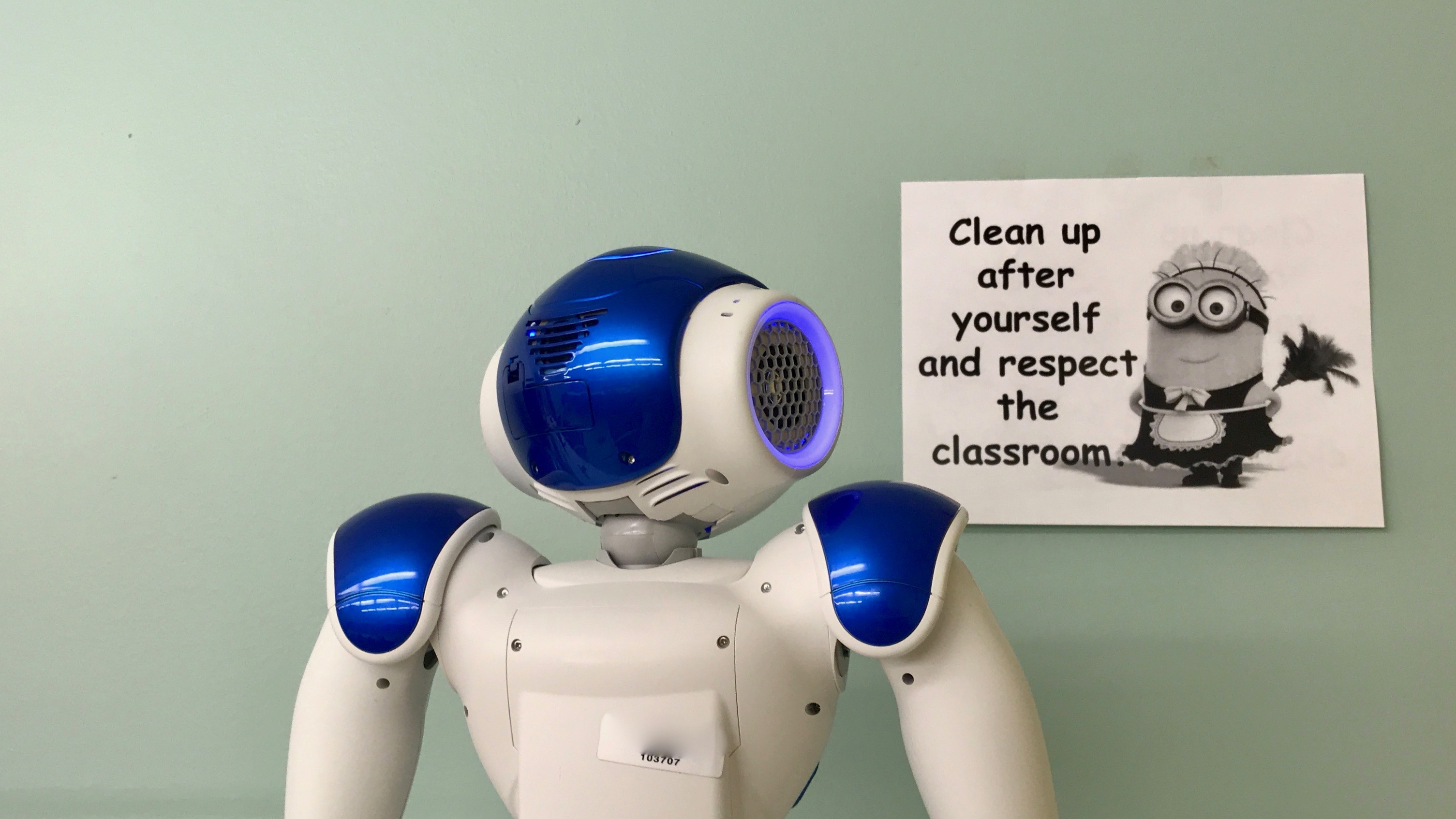}
\end{center}
\caption{The NAO robot learns to visually center the object, ie. sign, by moving its head.}
\label{fig:blue}
\end{figure}

\subsection{Experiential Robot Learning}
The principles of Experiential Robot Learning outlined in \citep{aly2017experiential} pose some constraints on how agents should be designed and trained. The process of Learning is considered lifelong and open-ended.  Thus, Agents must be capable of knowledge-acquisition, ideally autonomously or through minimal supervision. Without the prospect of explicit re-programming, the agent needs to adapt to the environment through other means such as interactive goal-setting or exploration, ie. curiosity. In addition, the method prescribes the use of Deep Learning techniques to implement the agents.

The main motivation behind using Experiential Robot Learning, or ERL, is that it is designed to tackle the limitations of hand-coding. If an agent encounters unknown conditions or adversity and it consistently needs to be programmed explicitly. This unscalable process becomes a bottleneck to agent development. The concern, thus, is to create algorithms capable of open-ended autonomous lifelong learning. Of course, there is no one general algorithm to do this, however. Rather, there needs to be a fusion of learning algorithms. We choose to adopt this method throughout this work as our main inspiration.

\subsection{Gradient-based Optimization}
The algorithms of Stochastic Gradient Descent and Backpropagation were invented decades ago \citep{Bottou:2016aa}\citep{SCHMIDHUBER201585}. There have been many variants that improve the performance of SGD such as the widely-adopted ADAM algorithm \citep{Kingma:2014aa}. Those algorithms gave rise to what is now dubbed Deep Learning. In effect, Deep Learning can generally be considered as gradient-based optimization of deep neural networks.

Gradient-based methods, however, pose constraints such as end-to-end differentiability of the model. When dealing with some tasks, for example Hard Attention \citep{Luong:2015aa}, this differentiability requirement can become a limitation. In addition, when one explores the field of Reinforcement Learning, the constraints are more pronounced. Agents suffer from lack of exploration and stagnation due to the natural behavior of SGD and uninformative gradient signals \citep{Khadka:2018aa}. There has been many techniques to get around those limitations in Computer Vision \citep{pmlr-v37-xuc15} and Reinforcement Learning \citep{pmlr-v48-mniha16}.

\subsection{Accelerated Deep Neuroevolution}
In this work, we propose amendments to  the Experiential Robot Learning method based on the aforementioned observations concerning gradient-based optimization. Primarily, we propose the usage of Deep Neuroevolution as an alternative optimization technique. 

Genetic algorithms, have been notoriously known for their convergence problem. There are multiple attempts to combat this, most notably is the NEAT algorithm \citep{Stanley_2002}, and its variants such as FS-NEAT \cite{Whiteson_2005}. However, those approaches rely on mutating the network structure alongside weights.

In addition, due to the large number of generations and populations within each generation, the number of physical implementations of Evolutionary Algorithms has thus far been limited \citep{Lahr_2017}. This is especially pronounced when taking into account that ERL requires online learning, though it does allow a bootstrapping process to precede.

In this context, we propose an Accelerated Neuroevolution, or ANv1, algorithm that improves upon a baseline algorithm in the game Flappy Bird. Our algorithm is engineered to display several desirable behaviors such as faster convergence, stability and adaptive exploration. We deemed ANv1 suited for physical implementation, and tested it on a NAO robot in a visuomotor learning task. The goal is to center an object in the robot's field of vision. A visual reference of the task is given in Figure \ref{fig:blue}.

In addition, we introduce a gradient-trained CNN that classifies the location of an object. The CNN is seamlessly integrated with an evolved network, to create a hybrid system of gradient and gradient-free trained networks. ANv1 performs remarkably in a physical, although simplified, environment.

\section{Experimental Outline} 
\label{methodology}

\subsection{Baseline Algorithm}
We chose the algorithm in \cite{Batchu_2017} as a baseline for our experiments. Briefly, the Batchu algorithm exposes a neural network to an evolutionary process through favorable breeding of higher ranking populations according to a fitness metric. In the Flappy Bird game, the fitness metric is the performance of the agent playing the game, ie. avoiding obstacles. 

The Batchu algorithm features a coarse notion of crossover, the offspring's networks are composed of entirely swapped layers of weights. That is, the a layer's weights, incoming or outgoing, are entirely inherited from one of the parents. A random mutation is then applied to the weights, with a selection probability of 0.15. 

The self-reported performance in of the algorithm is about 1 hour, ie. hundreds of generations, of training for convergence on a capable agent \citep{Batchu_2017}. This result corresponds to a fully-connected network with a hidden layer of 7 neurons, and a population size of 100 agents for each generation.

\subsection{Proposed Algorithm, ANv1}
\begin{table}[t!]
\caption{Outline of the proposed algorithm ANv1}
\label{proposed-algorithm}
\begin{center}
\begin{tabular}{ll}
\multicolumn{1}{c}{\bf STAGE}  &\multicolumn{1}{c}{\bf DESCRIPTION}
\\ \hline \\
Assemble new generation       	&- Assemble generation of N populations \\
						& \\
Perform the task             		&- Perform the assigned task, the populations experience \\
				      		&the environment either simultaneously or separately \\
						& \\
Evaluate Performance        	&- Evaluate the performance of each population \\
						&according to the fitness metric \\
						&- Determine the Winner as the population with the\\
			  			&highest score\\
             				 	&- Determine the Mutation Resistance rate based on the \\
					 	&calculated fitness metric of the Winner (either reset to \\
						&the original rate of 0.95 or lower by 0.05) \\
						& \\
Selective Breeding			&- Preserve the Winner \\
						&- Perform Genetic Crossover for the Royal Family \\
						&- Perform Genetic Crossover for the rest of the populations \\
						& \\
Random Mutations			&- Preserve the Winner \\
						&- Perform Mutation for all other populations using the \\
						&newly-determined Mutation Resistance rate \\
\end{tabular}
\end{center}
\end{table}

The proposed algorithm, Accelerated Neuroevolution v1 or ANv1, needs to be much more efficient if it will be used for physical implementation. This may come at a cost of reduced exploration behavior. We modified a few behaviors of the baseline algorithm as such. The algorithm outline is given in Table ~\ref{proposed-algorithm}.

First, in Genetic Crossover instead of taking an entire weights layer from a parent, the weights of the offspring are taken in a criss-cross, ie. checkered, pattern from both parents. There are two offsprings from this process, depending on which parent we start with. Thus, if the first weight belongs to parent 1 then the second weight will belong to parent 2, and so on. Our intuition was that this equal participation allows the inheritance of both good and bad genes from the parents. In addition, it may encourage sparsity in the model.

Second, we designed an adaptive mutation scheme to balance perturbation, ie. exploration, and stability. The scheme randomly mutates weights with a 5\% selection probability. We call the inverse of this selection probability, ie. 95\%, Mutation Resistance rate. The rate is dynamically adjusted by comparing generational entropy of the fitness metric. For each generation after the first, the fitness metric is examined and compared with its predecessor. If the score is found to lie within a pre-defined interval, eg. -5-10\%, then it is dubbed stagnant. If there is a stagnant generation, the Mutation Resistance is accumulatively lowered by 5\% for the upcoming generation. Each stagnant generation thus experiences a higher probability of mutations. If finally one generation's fitness metric is sufficiently higher or lower than the stagnation interval, then the Mutation Resistance is reset to 95\%. 

We wanted to balance exploration and convergence. The algorithm needed to be able to get out of local minima, and our intuition was that this behavior can help with that. Note that the magnitude of the mutation is constant, it is the mutation selection probability that changes adaptively.

Third, we made amendments to help increase convergence rates. In our experience the baseline algorithm did not exploit breakthroughs greedily enough. If for example one of the populations was able to demonstrate a better behavior than the others, its good genes are usually lost in the crossover and mutation process. Therefore we introduced a Winner and Royal Family concept. 

The Winner of each generation is the best-performing agent. This Winner is preserved to the following generation, without experiencing mutation. In addition, the Royal Family is the winner breeding, ie. performing genetic crossover, with itself, and undergoing mutations. The size of the Royal Family is predefined at the start of the algorithm. We typically choose a low number such as 4. Finally, the rest of the populations breeds with the Winner and undergoes mutations as normal. 

In this way, the Winner's good genes are dissipated to the entire upcoming generation, with various concentrations. The preservation of the Winner agent ensures this unique genetic makeup endures without mutation. The Royal Family population aims to improve the Winner's genetic makeup by exposing it to mutations. Breeding the rest of the population with the Winner rather than amongst themselves aims to create radically new genetic makeups based on the Winner's while allowing the inheritance of bad genes. We view bad genes as necessary to create new behaviors and encourage entropy within the genetic pool. In essence, with those choices we attempt to maximally capitalize on the good performance of any particular agent.

\subsection{Location Net}
We developed our algorithm described in the previous section to perform in the physical environment. The Flappy Bird game is simply a demonstration of its behavior in a simulated environment. The task we chose for the first implementation is the centering of a static object in the robot's field of vision. It is a stepping stone for more complex physical learning tasks such as real-time moving-object tracking.

We trained a CNN to classify an object's location in the robot's field of view. The network divides the visual field into a 3x3 grid, a class for each section. In addition, there's a Null class if there's no object detected in the image. Namely, the 10 classes are: Top Left, Top Center, Top Right, Left, Center, Right, Bottom Left, Bottom Center, Bottom Right and No Image. Location Net is designed to center any object regardless of what it is. The choice of using a sign, as shown in Figure \ref{fig:blue}, is only for convenience. The sign can be replaced with any other object.

\begin{figure}[t]
  \centering
  \begin{subfigure}[b]{0.2\linewidth}
    \includegraphics[width=\linewidth]{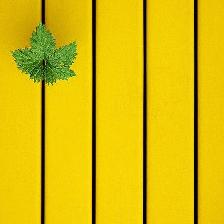}
     \caption{Top Left.}
  \end{subfigure}
  \begin{subfigure}[b]{0.2\linewidth}
    \includegraphics[width=\linewidth]{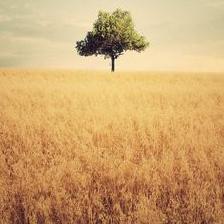}
    \caption{Top Center.}
  \end{subfigure}
  \begin{subfigure}[b]{0.2\linewidth}
    \includegraphics[width=\linewidth]{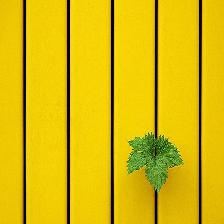}
    \caption{Bottom Right.}
  \end{subfigure}
  \begin{subfigure}[b]{0.2\linewidth}
    \includegraphics[width=\linewidth]{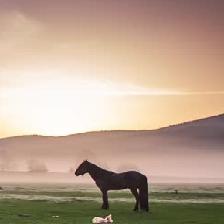}
    \caption{Bottom Center.}
  \end{subfigure}
  \caption{Training samples for different classes of the Location Net CNN. Note that (a) and (c) show the same object but in different locations, we call this an adversarial sample. The training set contains many such samples. The intuition in doing this, is that it encourages the network to focus more on the locations of objects rather than on the object itself.}
  \label{fig:locnet_samples}
\end{figure}

The network is based on a modified VGG16 architecture. Specifically, the fully-connected layers are shrunk to accommodate the relatively simple classification task. It is trained on a manually-assembled dataset of objects against a plain background in various locations in the 3x3 grid. In total, the training and validation sets contain over 1,500 images. The training and validation sets feature adversarial examples by having the exact same object and the exact same background but in different locations. Intuitively this encourages the classifier to recognize the locations rather than the objects in an image. Samples of the training and validation sets are given in Figure \ref{fig:locnet_samples}.

Location Net, or LocNet for short, was trained with the ADAM optimizer and written in Keras \citep{chollet2015keras} with a Tensorflow \citep{Abadi:2016aa} backend. It is a gradient-trained network.Training takes less than 10 minutes on an Nvidia Titan Xp GPU. The network achieves over 90\% accuracy for both the training and validation sets after 35 epochs. We deem this sufficient accuracy for the given task. Our task is neither time-sensitive nor action-critical, hence we can tolerate this error. The network can correct misclassifications through the redundancy of a video stream.

\subsection{Object Centering task}
In this task, the robot centers a static object in its field of vision by moving its head left or right. The robot's head is controlled by an evolved neural net. This network is seamlessly stacked under the Location Net CNN, described in the previous section. That is, the evolved net takes as input the output vector of the CNN and maps that to a command. The control net issues one of three commands: Move Left, No Movement, Move Right. Hence, we have a discrete control scheme, with a single Degree-of-Freedom, DOF.

Over the generations, the evolved control net has to figure the mapping of the classification to the correct control action. For example, if the CNN classifies the object to be on the 'Left', then the net has to issue a 'Move Left' command. The robot then moves shifts its Head Yaw angle to the Left by a fixed amount. Once the object is classified as 'Center', the net has to issue 'No Movement' commands to ensure stability.

This mapping is reflected by reward assignment. For every correct action the network's total reward is incremented by 100. At the end of the 5 episodes, the total reward reflects the network's performance. This total reward is our 'fitness metric', through which the network is evolved. Intuitively, this reward assignment is what guides the ANv1 algorithm's, ie. Robot's, behavior. If we want a different behavior, all we need is to change the reward assignment scheme.

\subsection{Setup}
In the following section we outline the experimental setup used to produce the presented results. In the first set of experiments two algorithms play the game of Flappy Bird on a desktop workstation. We use a Nvidia Titan Xp GPU to calculate and implement the populations, ie. neural networks. The networks are written in Keras and implemented in Tensorflow.

For the second set of experiments, we use a NAO robot to center a physical static object. The algorithm runs on a desktop workstation with an Nvidia Titan Xp GPU. The robot captures its current view using its front camera, and sends it to the workstation over a wireless local network. The CNN accepts the image as input then each control network, ie. population, gets to control the robot for 5 episodes. The goal is to center the object within those 5 episodes, and ensure stable convergence.

Each 'Move' command increments the current position of the Head Yaw joint angle by a fixed amount. This amount is small enough to allow fine control, and limit overshooting. It is empirically chosen, based on observation. Incrementation is implemented by querying the robot for the current joint angle, and adjusting it based on the neural network's choice. Finally, the new position is sent to the robot as a command over a wireless local network.

\section{Results}
\subsection{Flappy Bird game}
In this section we present the results of playing the Flappy Bird video game both with the baseline algorithm and our own algorithm. Note that the game has no end. That is, an agent capable of playing the game, can potentially go on indefinitely. This means there is no definite metric for convergence. For that reason, we qualitatively pronounce whether an agent is capable of playing the game based on our own observations. 

However, we understand that qualitative assessment is not a robust benchmark to compare performance. Thus, we report the generational score, normalized for population size, over the number of generations as a means to compare the algorithms. Normalized generational score is the aggregate score for the entire generation divided by the number of populations. This gives us an idea of how the entire generation performs rather than an individual outlier. In addition, we chose the number of generations, ie. rollouts, as measure instead of time, because time may vary across tasks and domains. The number of rollouts, on the other hand, reflects the performance of the algorithm.

As a reminder, one of the motivations of creating our own algorithm is that genetic algorithms are generally known to be unfit for implementation on physical agents. This is due to their slow convergence and impracticality considering the population size.

\begin{figure}[h]
\centering
\begin{subfigure}[b]{0.4\linewidth}
\begin{tikzpicture}
\begin{axis}[
          width=\linewidth,
          grid=major,
          grid style={dashed,gray!30},
	  ymin = 0,
	  xmin = 0, xmax = 750,
          xlabel=Generations,
          ylabel=Normalized Score,
          no marks
        ]
\addplot table[no marks, x=generation, y=longgen, col sep=comma] {long.csv}; 
\end{axis}
\end{tikzpicture}
\caption{750 Generations.}
\end{subfigure}
\begin{subfigure}[b]{0.4\linewidth}
\begin{tikzpicture}
\begin{axis}[
          width=\linewidth,
          grid=major,
          grid style={dashed,gray!30},
	  ymin = 0,
	  xmin = 0, xmax = 375,
          xlabel=Generations,
          ylabel=Normalized Score,
          no marks,
        ]
\addplot table[no marks, x=generation, y=mediumgen, col sep=comma] {med.csv}; 
\end{axis}
\end{tikzpicture}
\caption{375 Generations.}
\end{subfigure}
\begin{subfigure}[b]{0.5\linewidth}
\begin{tikzpicture}
\begin{axis}[
          width=\linewidth,
          grid=major,
          grid style={dashed,gray!30},
	  ymin = 0,
	  xmin = 0, xmax = 100,
          xlabel=Generations,
          ylabel=Normalized Score,
        ]
\addplot table[x=generation, y=shortgen1, col sep=comma] {short.csv}; 
\addplot table[x=generation, y=shortgen2, col sep=comma] {short.csv}; 
\legend{Run 1, Run 2}
\end{axis}
\end{tikzpicture}
\caption{100 Generations.}
\end{subfigure}
\caption{Different, individual rollouts of varying lengths of the baseline algorithm playing Flappy Bird. The chart reflects poor performance due to the large number of generations necessary to achieve learning as reflected by a higher score. The algorithm fails to capitalize on breakthrough generations by relapsing to poor performance.}
\label{fig:baseline}
\end{figure}
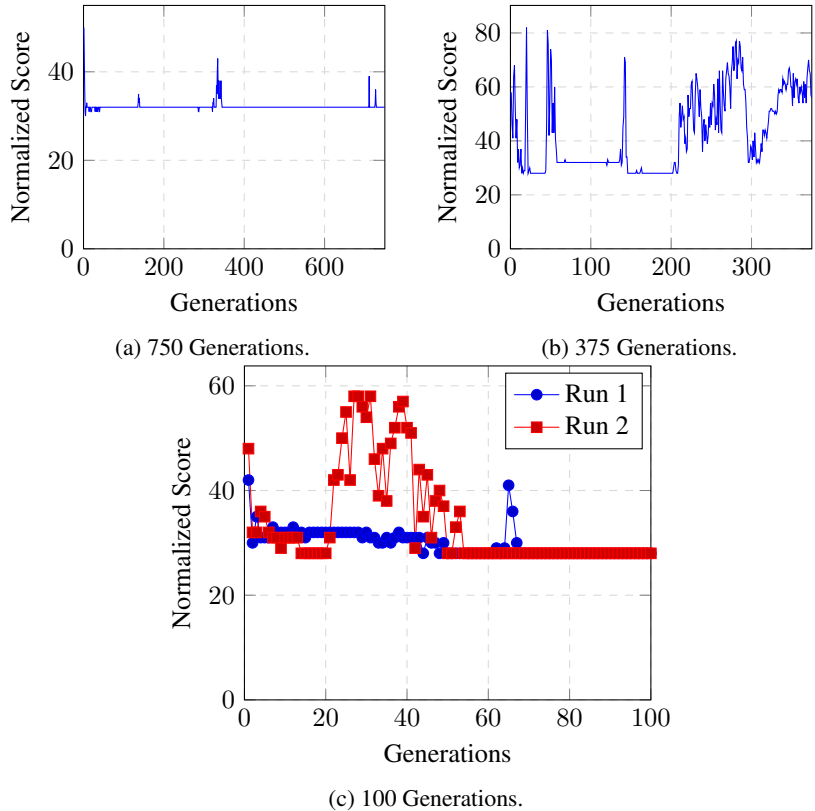

The baseline algorithm described in Section \ref{methodology} is reported to train agents capable of playing the game after roughly an hour of training, ie. over a thousand generations. The algorithm operates on a single-hidden layer network of 7 neurons. It takes as input a tuple of 3 numbers representing the distance to the nearest pipe, the gap height and the height of the agent. 

The self-reported results as well as those reported in Figure \ref{fig:baseline} for the baseline algorithm are for a population size of 50. That is, the algorithm is training 50 populations every generation. As the number of agents increase the chance that a population has good genes increases. This in turn leads to quicker convergence, intuitively. However, requiring a large number of populations for convergence renders an algorithm infeasible for physical implementation.

In Figure \ref{fig:baseline} we present the results of using the Batchu algorithm over rollouts of varied lengths. It performed rather poorly. Notably, convergence is extremely slow, and the algorithm fails to capitalize on breakthrough generations. Ideally, we want learnings to compound over generations for quick convergence. In addition, we also see how sensitive it is to initialization. Each plot in the figure is a separate run. Performance varies drastically between (a) and (b) for example. From Run 2 in (c) we highlight how performance regresses eventually despite a promising start.

The algorithm we present, and described in Section \ref{methodology}, operates on a single-hidden layer network of 50 neurons. We increase the number of neurons by 7x from the baseline algorithm. In practical implementations, problems are not going to be as simple as Flappy Bird. We wanted to test how our algorithm responds to a larger network. We use the same population size of 50.

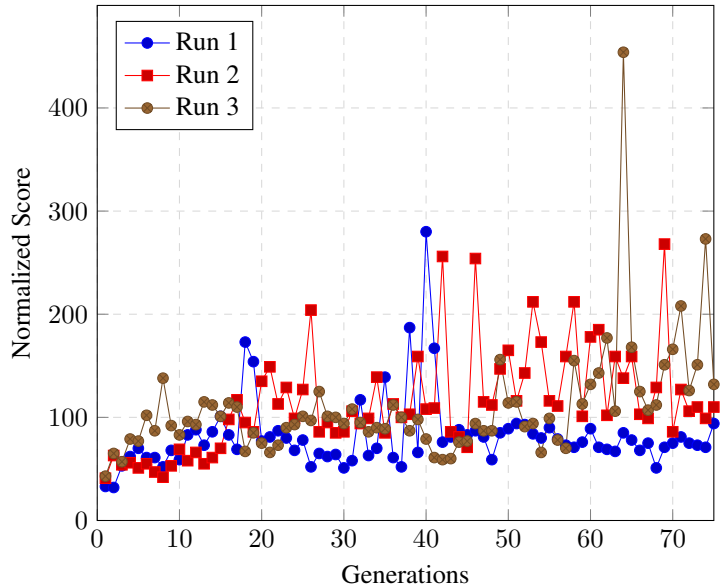
\begin{figure}[h]
\centering
\begin{tikzpicture}
\begin{axis}[
          width=0.7\linewidth,
          grid=major,
          grid style={dashed,gray!30},
	  ymin = 0,
	  xmin = 0, xmax = 75,
          xlabel=Generations,
          ylabel=Normalized Score,
          legend pos = north west
        ]
\addplot table[x=generation, y=score3, col sep=comma] {myalgo.csv};
\addplot table[x=generation, y=score4, col sep=comma] {myalgo.csv};
\addplot table[x=generation, y=score5, col sep=comma] {myalgo.csv};
\legend{Run 1, Run 2, Run 3}
\end{axis}
\end{tikzpicture}
\caption{Performance of the ANv1 algorithm. The algorithm performs at least 1.7x better than baseline, over a much lower number of generations, while evolving a network that is 7x larger. The algorithm generally capitalizes on breakthrough generations, even if subsequent generations are not as well-performing.}
\label{fig:myalgo}
\end{figure}

The performance of the ANv1 algorithm is presented in Figure \ref{fig:myalgo}. The algorithm is reported over only 75 generations. As the agents get better at playing the game, it takes longer for them to commit a mistake. Training thus becomes a diminishing returns factor of improvement. As we can see, the algorithm performs much better than baseline on average, an improvement of between 1.7x and 7.1x. Of course, those improvements arrive in a much shorter span of generations.

We can see behaviors in Figure \ref{fig:myalgo} such as getting out of local minima by the accelerated compounding of score in the early stages. The dynamism of ANv1 is also highlighted by how the score doesn't plateau for more than a handful of generations. This behavior was engineered into the algorithm in order to achieve higher scores through exploration. We also note how the algorithm almost always performs better after a breakthrough generation, than pre-breakthrough. It shows that the algorithm capitalizes on breakthrough generations. This also an intended behaviors that has been engineered. Stability and robustness can be highlighted by how the algorithm does not regress or backtrack to poor, initial performance. This is subtly different from capitalization on breakthrough generations.

\subsection{Object Centering task}
This is arguably the most important section of the paper. We deemed the ANv1 algorithm ready for implementation on a physical system. After all, it was our main motivation in pursuing this thread of inquiry. We decided to shrink the population size to only 15 in each generation. This number lowers the exploration potential of the algorithm, but makes it far more practical. The network is still a single-hidden layer network, but of 10 neurons. This width is lower than the 50 neurons used in the earlier section. We chose this as the task is relatively simple and shouldn't require a network of larger representational capacity.

\begin{figure}[h]
\centering
\begin{tikzpicture}
\begin{axis}[
          width=0.4\linewidth,
          grid=major,
          grid style={dashed,gray!30},
	  ymin = 0,
	  xmin = 1, xmax = 9,
          xlabel=Experiment,
          ylabel=Generations to convergence,
          legend pos = north west
        ]
\addplot table[x=experiment number, y=number of generations, col sep=comma] {robot.csv};
\end{axis}
\end{tikzpicture}
\caption{Performance of the ANv1 algorithm in the static object centering task. We ran 9 experiments, three for each possible starting location, Left, Right and Center. The algorithm terminates when 7 out of the 15 populations collect maximum rewards. The algorithm converges in less than 5 generations, with one exception.}
\label{fig:robot}
\end{figure}
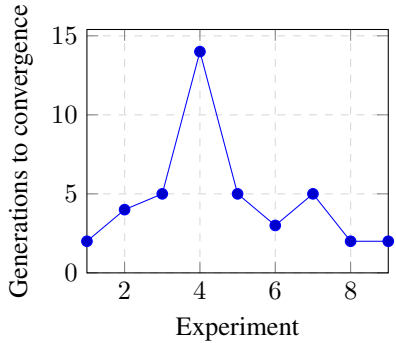

The task is described in Section \ref{methodology}. To reiterate, the robot is required to 'center' a static object by moving its head left and right. For each generation, ie. iteration, there are 15 populations. Each population controls the robot for 5 episodes. To ensure robustness, we start the robot in 3 different locations. Specifically, we place the robot to the Left, Center and Right, of the object. For each location we carry our experiment 3 times. This is a total of 9 experiments.

The algorithm stops when 7 of the 15 populations have received maximum rewards. That is, 7 or more populations successfully map the CNN classifications to the correct commands for all the 5 episodes, ie. optimal agents. Once this threshold is achieved or passed, the algorithm terminates.

We have found that the ANv1 algorithm successfully navigates the task 9 out of 9 times. Each experiment generally took 10 minutes or less, with one exception. The results are shown in Figure \ref{fig:robot}. Experiment 4 in Figure \ref{fig:robot} took 14 generations to convergence while the other two experiments in its class, 5 and 6, took 5 and 3 generations respectively. We assign this behavior to sensitivity to initialization. To a lesser extent, conservative mutations may also be a contributor. In ANv1, the initial mutation resistance rate, as outlined in Section \ref{methodology}, is 95\%; and it decreases only in 5\% decrements per generation. In general, we are satisfied that our choice of running 3 experiments per location has been rewarded in revealing this behavior.

\section{Conclusions}
Motivated and guided by the Experiential Robot Learning, ERL for short, method, we proposed the Accelerated Neuroevolution v1, ANv1 for short, algorithm. We tested the algorithm in two scenarios. The first experimental set is in a simulated environment, playing the game of Flappy Bird. We found that the algorithm performs much better, 1.7-7x, than a baseline algorithm, while controlling a network that is 7x larger. We observed encouraging behavior such as capitalization on breakthrough generations, ability to get out of local minima, high dynamism while maintaining robustness and stability. 

The second experimental set took place in the physical environment on a NAO robot. The goal was to center an object in the robot's field of vision. We developed a CNN called Location Net to classify where the object is. We found that the algorithm generally converges in a relatively low number of generations, 5 or less. We did observe, however, that the algorithm is sensitive to initialization. This is noted in the case of experiment 4 in Figure \ref{fig:robot} where the initialization was likely adversarial.

While not reported here, we also observed how the algorithm makes heavy use of the adaptive mutation resistance rate. Whenever it was 'stuck' or plateauing, the algorithm would decrease the resistance rate until it got out of the flat region. We observed this for both the simulated and the physical experimental sets.

We recognize the limitations of our work, as well. The algorithm needs to be benchmarked on larger, more complex tasks. In addition, more realistic tasks will require evolving neural networks of greater representational capacities, ie. deeper, wider and more varied. However, a single DOF task is not unheard of in the domains of Robotics and Reinforcement Learning. Furthermore, given that this is an initial step, we believe it is a significant proposition and validation.

For future iterations, we aim to implement adaptive mutation magnitudes. While the mutation resistance rate is adaptive, the magnitude of the mutations is static. We think this will become useful when evolving more complex networks, whose behavior needs to be fine-tuned. 

In conclusion, we presented a first foray into Accelerated Neuroevolution with a simulated and a physical learning experiment. The algorithm performed well in the assigned tasks, and demonstrated encouraging behaviors. We like how the integration between a gradient-trained network, ie. LocNet, was seamlessly integrated with an evolved network, ie. the control net. We see this sort of integration proliferating into more applications that prove that a hybrid approach is needed.

\bibliography{draft}

\end{document}